# Innovative Deep Learning Architecture for Enhanced Altered Fingerprint Recognition

Dana A Abdullah [1], Dana Rasul Hamad [3], Bishar Rasheed Ibrahim[4], Sirwan Abdulwahid Aula[3], Aso Khaleel Ameen[2], Sabat Salih Hamadamin[3].

[1] Faculty of Engineering and Computer Science, Qaiwan International University, Sulaymaniyah, KRG, Iraq.

[2] Department of Computer Science, College of Science, Knowledge University, Erbil 44001, Iraq.

[3] Computer Science Department, Faculty of Science, Soran University, Soran, Erbil, KRG, Iraq.

[4] Department of Computer Networks and Information Security, Technical College of Informatics – Akre, Duhok Polytechnic University, Duhok 42006, Iraq

## Abstract:

Altered fingerprint recognition (AFR) poses a significant challenge to biometric security or biometric verification systems to particular applications like border control, forensic or fiscal admission. People can purposefully alter finger print patterns as a way to evade detection and therefore it is crucial to have meaningful recognition of altered prints. The paper presents DeepAFRNet, a deep learning-based recognition model with the objective of matching and recognizing distorted samples of fingerprints. The suggested model combines VGG16 convolutional neural network to extract high-dimensional features and cosine similarity to compare fingerprint embeddings. The SOCOFing Real-Altered Subset was used, and each fingerprint image had three degrees of alterations Easy, Medium, and Hard. DeepAFRNet reported high levels of recognition accuracy (96.7%, 98.76%, and 99.54%) in the three respective categories when the thresholds were strict. A detailed threshold sensitivity analysis revealed that lowering the threshold (from 0.92 to 0.72) significantly degraded performance dropping accuracy to 7.86%, 27.05%, and 29.51%, respectively highlighting the importance of threshold selection in biometric systems. The model's robustness across different alteration levels and its adaptability to threshold configurations address several limitations of previous approaches, including reliance on synthetic data, lack of verification mechanisms, and limited feature representation. By using real altered fingerprint samples and providing detailed performance metrics per difficulty level, DeepAFRNet offers a comprehensive solution for AFR tasks. The findings emphasize its potential deployment in real-world applications where both security and recognition resilience are critical.

# 1. Introduction

## 1.1. Research background

AFR has emerged as a critical issue for maintaining the integrity of biometric identification systems used in security, personal devices, border control, and law enforcement. Alterations in fingerprint patterns can stem from both natural and deliberate sources [14]. Natural factors include injuries such as cuts, burns, or scars, as well as the effects of ageing, which subtly modify skin texture and disrupt the clarity of fingerprint patterns [10]. On the other hand, deliberate alterations involve methods like plastic surgery, abrasions, or chemical treatments, typically used by individuals seeking to evade detection in high-security or criminal scenarios [3] [12]. These alterations challenge conventional recognition algorithms, which are primarily designed to detect consistent, unmodified ridge patterns [29]. The inability of traditional systems to effectively identify altered fingerprints presents a significant vulnerability, potentially allowing individuals to bypass security protocols [27] [19]. There are some statistical facts showing fingerprint alteration being under focused, for example a report based on the FBI Integrated Automated Fingerprint Identification System (IAFIS) database found that the agents detected 412 fingerprints records on which there were indications that revealed deliberate alteration [21]. The states where such instances have been the most common were Massachusetts, New York, Texas, California and Arizona which shows a regional tendency of trying to cheat biometric identification systems. This underscores an emerging issue to the police and security forces in finding and managing modified fingerprints in national ID databases [24]. A study [11] conducted by Michigan State University utilized an academic dataset comprising 4,815 altered fingerprint samples collected from 270 individuals. The evaluation of detection performance revealed a True Detection Rate (TDR) of 99.24% at a False Detection Rate (FDR) of 2%, demonstrating the high effectiveness of advanced recognition algorithms in accurately identifying altered fingerprints.

Fingerprint recognition requires minimal effort from the users [6]. However, there are some barriers to addressing these challenges [9]. The integration of artificial intelligence (AI) and deep learning has proven transformative [8]. Deep learning models, especially CNNs, excel at automatically extracting complex, high-dimensional features from images, making them well-suited for analysing variations in fingerprint patterns [1] [20]. AI-powered systems can better differentiate between authentic changes in fingerprints due to natural causes and deliberate modifications aimed at evading detection [30][16] [19].

The proposed solution, DeepAFRNet represents a significant advancement in AFR technology. DeepAFRNet leverages the capabilities of CNNs to learn and extract intricate, distinguishing features from fingerprint data, capturing both subtle and overt alterations. Unlike traditional algorithms that struggle with variability, DeepAFRNet's architecture is specifically designed to handle complex modifications introduced by injuries, ageing, or deliberate tampering. DeepAFRNet incorporates state-of-the-art feature extraction techniques that build upon the success of pretrained models like VGG16, known for their robustness in biometric recognition tasks. By tailoring these techniques to target both real and altered fingerprints, the model enhances recognition accuracy and reliability. It utilises similarity metrics, such as cosine similarity, to quantify the closeness of the extracted feature embeddings, facilitating more precise differentiation between genuine and altered fingerprints.

The outcome is a deep learning model that significantly improves the performance of AFR systems. DeepAFRNet not only boosts accuracy but also enhances scalability, ensuring that biometric systems can handle a wide range of fingerprint variations. This approach mitigates the security risks associated with fingerprint manipulation and provides a robust defence against spoofing attempts. By embedding deep learning within the fingerprint recognition framework, DeepAFRNet addresses the limitations of conventional methods, contributing to the security, reliability, and trustworthiness of biometric systems.

## 1.2. Contribution

- Proposal of DeepAFRNet for Altered Fingerprint Recognition (AFR): This work proposes DeepAFRNet, a robust framework for matching and recognising real and altered fingerprints, leveraging deep feature extraction and similarity measurement. It enhances fingerprint recognition accuracy by effectively distinguishing between real and altered patterns.
- Utilisation of Pre-Trained CNN Models: Integration of a pre-trained (VGG16) model for feature extraction, showcasing its robustness and adaptability in capturing intricate fingerprint features [25].
- Effective Feature Comparison Using Cosine Similarity: This work introduces cosine similarity metrics to compare feature embeddings, enabling precise recognition and matching of altered fingerprints with their corresponding real counterparts by identifying the highest similarity.
- Advancing Biometric Security: Highlighting the transformative role of deep learning in biometric systems, DeepAFRNet sets a new standard for reliability, robustness, and security in AFR.
- Enhancement of AFR System Scalability and Security: Contribution to the scalability of AFR systems by developing a solution that is resilient to diverse alterations, including natural and deliberate modifications. This strengthens the defence against fingerprint spoofing and manipulation.

## 2. Related Work

This section discusses related works that are closely aligned with this study, particularly in the area of recognising different types of altered fingerprints compared to real ones. Due to the limited research available in this domain, three key works have been selected for review. Each of these studies is discussed in detail in the following paragraphs.

[32]This paper presents a method for detecting and localising fingerprint alterations, which are intentional changes to fingerprint patterns to evade Automated Fingerprint Identification Systems (AFIS). The main contributions are: (i) developing and training Convolutional Neural Networks (CNNs) on fingerprint images and minutiae-centred patches to identify and locate altered regions, and (ii) training a Generative Adversarial Network (GAN) to generate synthetic altered fingerprints resembling real ones, addressing the scarcity of altered fingerprint data. Using a dataset of 4,815 altered fingerprints from 270 subjects, the method achieves a True Detection Rate (TDR) of 99.24% with a False Detection Rate (FDR) of 2%, surpassing previous results. The model, code, and synthetic dataset will be open-sourced.

[18] This paper proposes an effective method for classifying fingerprint images as authentic or altered, addressing the issue of criminals and hackers modifying fingerprints. The method enhances fingerprint classification accuracy by utilising texture features, specifically the Histogram of Oriented Gradients (HOG) and Segmentation-based Feature Texture Analysis (SFTA), and employs Discriminant Analysis (DCA) and Gaussian Discriminant Analysis (GDA) as classifiers. The method is tested using the Sokoto Coventry Fingerprint Dataset (SOCOFing), which contains 6,000 fingerprint images with different degrees of alteration. The proposed approach achieved a high classification accuracy of 99%, outperforming existing methods in both feature dimension and classification accuracy.

[17] This study addresses the issue of altered fingerprints, which are intentionally modified by individuals to evade identification by law enforcement. Using a deep neural network approach based on the Inception-v3 architecture, the paper proposes a method for detecting altered fingerprints, classifying types of alterations, and identifying gender, hand, and finger associated with the fingerprints. The study also generates activation maps to visualise areas of the fingerprint that the network uses for detecting alterations. The method achieves high accuracy on the SOCOFing dataset,

with 98.21% for fakeness detection, 98.46% for alteration classification, 92.52% for gender identification, 97.53% for hand classification, and 92.18% for finger classification.

## 3. Methodology

This section outlines the dataset collection, pre-processing, and the proposed model design. A dataset was collected from Kaggle sources, ensuring representation across different attributes. Pre-processing included label verification to prepare the data for model training, format conversion and data augmentation (e.g., rotation, flipping, and noise addition). The proposed model, DeepAfrNet, was designed using advanced deep learning techniques to extract meaningful features and achieve optimal performance. Its architecture and components are detailed, highlighting the rationale behind its design and its suitability for the task

### 3.1. Dataset

Finding datasets that include both real and altered fingerprints is challenging due to issues such as limited diversity in alterations, varying image quality, legal and ethical constraints, and dataset size. The SOCOFing Real-Altered Subset dataset from Kaggle is sourced [26] addresses these challenges effectively by providing a well-curated collection of fingerprint images with both genuine and altered samples. Specifically, it features 4 subjects with 3 types of (easy, medium and hard) alterations, offering a realistic and comprehensive resource for research. Kaggle's reputation for reliable datasets and clear usage guidelines ensures that the SOCOFing dataset is of high quality and legally compliant. By leveraging this dataset, researchers can develop and evaluate fingerprint recognition systems that are robust and effective in handling real-world variations and alterations, advancing the field with more accurate and resilient technologies. The above details about the employed dataset have been presented in Figure 1.

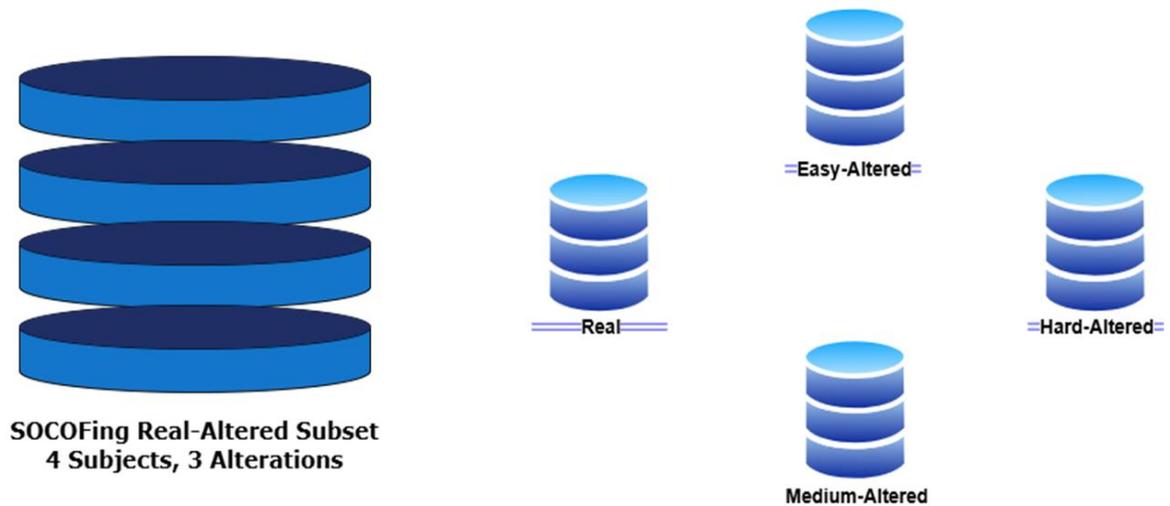

*Figure 1. SOCOFing dataset presentation*

In this study, we utilised the SOCOFing Real-Altered Subset dataset, which includes data from four subjects and three types of alterations. The dataset is divided into four categories: Real, Easy-Altered, Medium-Altered, and Hard-Altered. It comprises a total of 301 fingerprint images.

The primary reason for selecting this dataset is its recent update on Kaggle, ensuring the data's relevance and accessibility. Additionally, the dataset was chosen because the gender of the subjects and the specific type of finger are not considered critical factors for the analysis conducted in this research; the division rate of the dataset instances is illustrated in Figure 2.

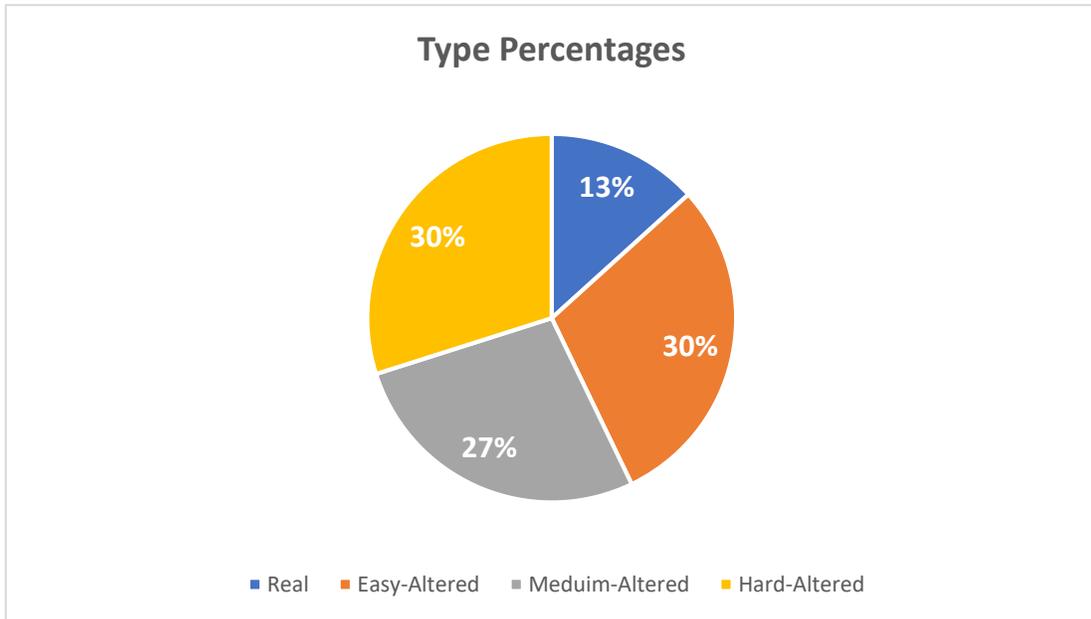

*Figure 2. Dataset distribution percentages.*

## 3.2. Processing

### 3.2.1 Relabelling

To enhance model performance and ensure consistency, relabelling was performed on the SOCOFing dataset. The original filenames of the fingerprint images were lengthy and contained special characters, which could potentially disrupt the model's processing and performance. [5] state that shortening the file characteristics of data handlers ensuring the quality of the datasets. By simplifying the filenames and removing special characters, the dataset was standardised, making it easier to manage and utilise in machine learning workflows. This step ensures that the image names do not adversely affect the model's ability to train and recognise patterns effectively.

### 3.2.2 Format Conversion

Initially, the fingerprint images in the dataset were stored in BMP (Bitmap) format. BMP files are uncompressed and can be quite large, which might lead to inefficiencies in data processing and increased storage requirements. To address this, all images were converted to PNG (Portable Network Graphics) format. PNG is a compressed format that retains high-quality image data while reducing file size, making it more efficient for processing and storage. This conversion helps in optimising the dataset for faster training times and reduced resource consumption [28].

### 3.2.3 Data Augmentation

Data augmentation was applied to both real and altered fingerprints to expand the dataset and enhance model training. This technique involves generating additional variations of the existing images through transformations such as rotation, scaling, flipping, and colour adjustments. The purpose of data augmentation is to artificially increase the size of the dataset, providing a more diverse set of training examples [13][4]. This helps improve the robustness of the fingerprint recognition models by exposing them to a wider range of scenarios and variations [7], thus enhancing their ability to generalise and perform well on unseen data. Figure 3 shows the dataset pre-processing procedures.

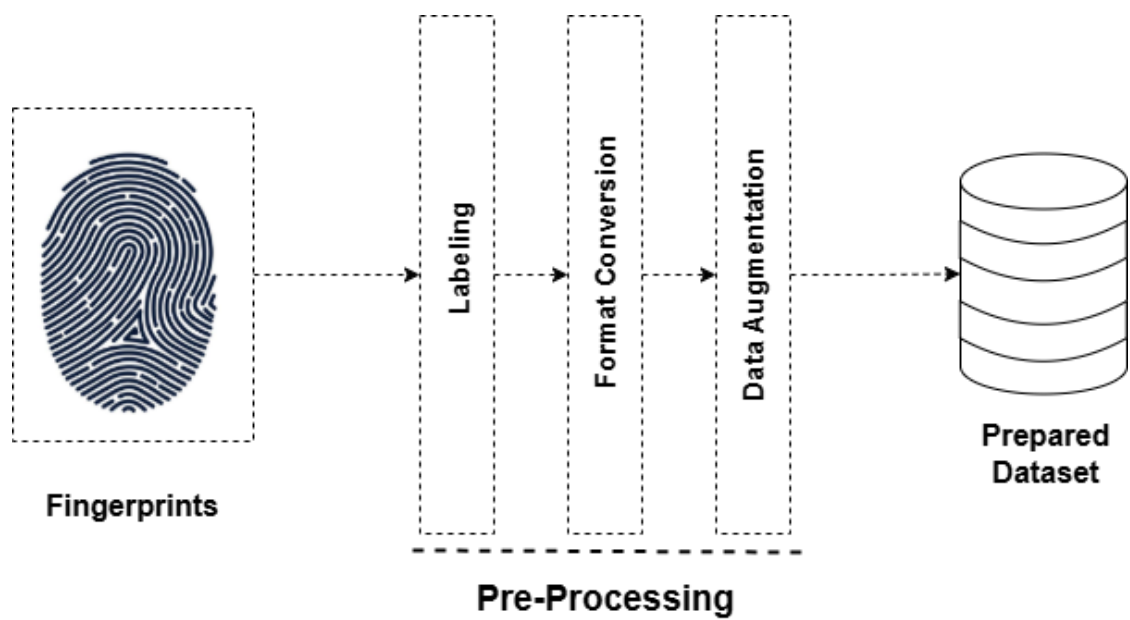

*Figure 3. Dataset pre-processing*

### 3.3. Proposed Model

In this research, the proposed model, DeepAfrNet, is discussed. This model has been specifically designed for recognising and matching real and altered fingerprints. It comprises two main components: the first is responsible for feature extraction from the input fingerprints, and the second calculates the percentage of similarity between the extracted features of real and altered fingerprints.

### 3.3.1 Feature Extraction with VGG16

For feature extraction, the pre-trained deep learning model VGG16 is utilised. General CNNs were not utilised as they did not perform well in capturing the complex features needed for recognising between real and altered fingerprints. VGG16 is a deep architecture, which includes 16 layers with small 3x3 convolutional filters, excels at identifying intricate patterns and details in fingerprint images [23]. The model consists of multiple convolutional layers followed by fully connected layers, which transform the input image into a high-dimensional feature vector. In the context of fingerprint recognition, VGG16 extracts unique and robust features from both real and altered fingerprints, such as ridge patterns, minutiae points, and textural characteristics [15]. These feature vectors provide a compact representation of the fingerprints, facilitating effective comparison.

### 3.3.2 Cosine Similarity Matching:

Feature extraction is performed using the pre-trained VGG16 model, which processes the fingerprint images and extracts key features from them. Once the features are extracted, cosine similarity is employed to compare the feature vectors of both real and altered fingerprints. This method is essential for determining if the fingerprints, one of which may have been altered, are from the same individual. Unlike traditional pixel-wise comparisons, which can be sensitive to slight distortions or variations in fingerprint patterns, cosine similarity focuses on the alignment of features, making it more robust to such changes.

For each fingerprint, the feature extraction process generates a high-dimensional vector that represents the fingerprint's unique characteristics. Cosine similarity is then calculated by comparing the feature vector of the real fingerprint with that of the altered fingerprint. The cosine similarity metric measures the cosine of the angle between these two vectors in multi-dimensional space, providing a value that indicates how closely related the two feature sets are. This value is independent of the magnitude of the vectors, meaning it focuses purely on their direction, which makes it an effective measure of similarity.

Once the cosine similarity score is computed, it is compared against a pre-defined threshold. If the score exceeds this threshold, the two fingerprints are considered to be a match, indicating that the altered fingerprint likely belongs to the same individual as the real fingerprint. The matched fingerprints are then printed and displayed for further verification. This method allows for highly accurate and efficient identification of individuals, even in cases where there are variations in the fingerprint patterns [31], as it emphasises the alignment of features rather than raw pixel values. Mathematically, cosine similarity is calculated as:

Cosine similarity between two vectors AAA and BBB is mathematically defined as shown in Equation 1 which is cited in [2].

$$Similarity\ (A, B) = \cos(\theta) = \frac{A.B}{\|A\|\|B\|} = \frac{\sum_{i=1}^{n} A_i B_i}{\sqrt{\sum_{i=1}^{n} A_i^2} \sqrt{\sum_{i=1}^{n} B_i^2}} \qquad \text{Equation 1}$$

Where A and B are the feature vectors of the two fingerprints being compared. A.B is the dot product of the two vectors. ‖A‖ is the magnitude (or norm) of vector A, calculated as Equation 2, and ‖B‖ is the magnitude (or norm) of vector B which is calculated in Equation 3.

$$\|A\| - \sqrt{\sum_{i-1}^{n} A \cdot B \sqrt{\sum_{i-1}^{n} A^2}} \qquad \text{Equation 2}$$

$$\|B\| - \sqrt{\sum_{i-1}^{n} A \cdot B \sqrt{\sum_{i-1}^{n} B^2}} \qquad \text{Equation 3}$$

Definitions in the Context of Fingerprint Recognition:

1. A (Real Fingerprint Vectors): Represents the vectorised features extracted from a real fingerprint image. These features could be minutiae points, ridge patterns, or other numerical representations derived from the fingerprint.
2. B (Altered Fingerprint Vectors): Represents the vectorised features extracted from an altered or tampered fingerprint image (e.g., a distorted or modified version of the original fingerprint).
3. Cosine Similarity: Measures the cosine of the angle (θ\theta) between the two vectors AA and BB. A similarity value closer to 1 indicates that the vectors are highly similar [22], implying that the altered fingerprint closely matches the real fingerprint.
4. Dot Product (A·BA \cdot B): Represents the sum of the products of corresponding elements in AA and BB, which quantifies their alignment.
5. Magnitude (‖A‖\|A\| and ‖B‖\|B\|): Represents the length of the vectors AA and BB. It ensures the similarity measurement is normalised to handle vector size differences.

The result of the equation is a value between −1-1−1 and 111:

- **1**: Perfect similarity, meaning the vectors point in the same direction.
- **0**: No similarity, meaning the vectors are orthogonal.
- **-1**: Perfect dissimilarity, meaning the vectors point in opposite directions.

### 3.4. Model Training and Testing

This section highlights the training and testing phases. After the model and pre-processing steps are set up using the collected data, the dataset will be split into training and testing portions. The dataset includes four categories: real, easy-altered, medium-altered, and hard-altered. More details about these categories will be explained in their respective sections.

### 3.5. Dataset Split:

In the "Dataset Splitting" section, the human fingerprint dataset, which includes four categories (real, easy-altered, medium-altered, and hard-altered), is separated into training and testing subsets. Each category is divided with 80% allocated for training and 20% for testing, using the train_test_split function from the scikit-learn library. This method ensures that the model learns from a well-rounded portion of the data while its performance is evaluated on a separate, unseen set.

### 3.6. Batch Size:

The batch size for training is set to 32, meaning the model processes 32 samples at a time before updating its weights. Using a smaller batch size, like 32, helps manage memory usage during training, though it requires more frequent updates to the model's parameters. In contrast, larger batch sizes reduce the number of updates needed but demand significantly more memory; the whole methodology is presented in Figure 4.

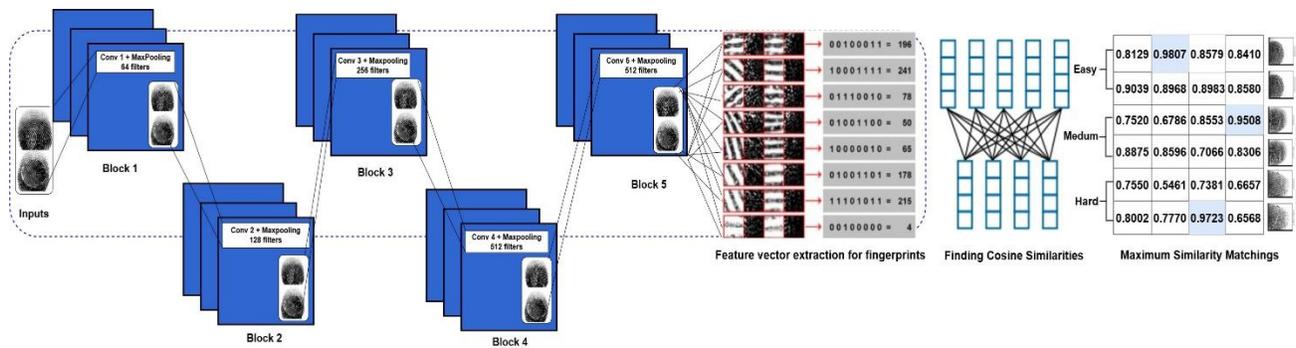

*Figure 4. Study process methodology stages.*

### 4. Results

After preparing the dataset and developing the DeepAFRNet model, the training phase utilised real fingerprint data to optimise the model's performance. Following this, the model was tested using altered fingerprint samples classified into three difficulty levels: Easy, Medium, and Hard. These levels were determined by the extent of modifications made to the fingerprints, representing different levels of complexity in matching altered fingerprints to their original versions. The performance of the DeepAFRNet model varied across these categories, underscoring its sensitivity to the degree of alteration in the input data.

The outputs of the model will be discussed as following:

## 4.1. Easy-Altered Recognition

The initial evaluation of the proposed DeepAFRNet model was conducted using the Easy-Altered fingerprint dataset, yielding a promising baseline accuracy of 92%. This high performance highlights the model's capability to effectively extract fingerprint features using the VGG16 convolutional base and compare them through cosine similarity, a robust similarity metric that is invariant to scale. The extracted feature vectors from real and easy-altered fingerprints (referred to as *real-vectors* and *easy-altered-vectors*) were compared in a high-dimensional space. High cosine similarity values indicated matched fingerprint pairs, whereas lower scores indicated non-matches.

To illustrate DeepAFRNet's robustness in detecting subtle alterations, a similarity-based evaluation was performed. In Table 1, a representative set of comparisons is shown, where one pair is correctly identified as a match with a high similarity score of 0.9808, while the rest were rejected due to insufficient similarity. These outcomes confirm the model's fine-grained sensitivity to alterations, even at the minimal level.

*Table 1. Easy-altered fingerprint match status of some samples.*

| Match Status | Real Fingerprint | Easy-Altered Fingerprint | Similarity |
|---|---|---|---|
| Matched | Real/6.png | Easy_Altered/18.png | 0.9808 |
| Not Matched | Real/6.png | Easy_Altered/19.png | 0.8636 |
| Not Matched | Real/6.png | Easy_Altered/20.png | 0.8129 |
| Not Matched | Real/6.png | Easy_Altered/21.png | 0.8579 |
| Not Matched | Real/6.png | Easy_Altered/61.png | 0.7517 |
| Not Matched | Real/6.png | Easy_Altered/62.png | 0.7257 |
| Not Matched | Real/6.png | Easy_Altered/63.png | 0.7742 |
| Not Matched | Real/6.png | Easy_Altered/64.png | 0.7176 |
| Not Matched | Real/6.png | Easy_Altered/65.png | 0.7925 |
| Not Matched | Real/6.png | Easy_Altered/66.png | 0.7914 |
| Not Matched | Real/6.png | Easy_Altered/67.png | 0.8339 |
| Not Matched | Real/6.png | Easy_Altered/68.png | 0.8136 |
| Not Matched | Real/6.png | Easy_Altered/69.png | 0.8477 |
| Not Matched | Real/6.png | Easy_Altered/70.png | 0.8398 |

From the above table, it is evident that DeepAFRNet successfully recognises high-similarity pairs, as shown by the matched pair with a cosine similarity score of 0.9808. This pair is visualised in Figure 5, which clearly illustrates the close resemblance between the real and the easy-altered fingerprint, despite minor distortions.

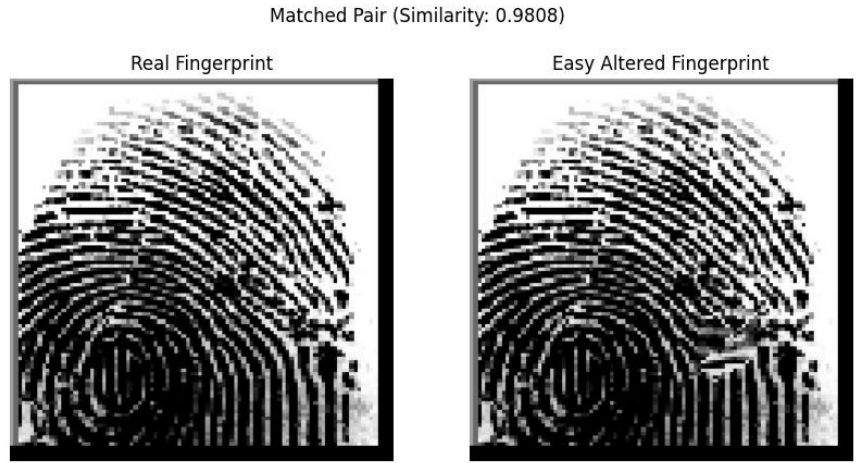

*Figure 5. Real and easy-altered fingerprint detection*

Further validation was conducted using a complete Easy-Altered test set to assess how model performance varies across different threshold levels. Specifically, we examined three thresholds: 0.92, 0.82, and 0.72, representing strict, moderate, and relaxed decision boundaries, respectively. This allowed for evaluating the impact of threshold tuning on accuracy, false match rate, and computational cost. At the highest threshold (0.92), the model correctly classified most pairs as non-matches, achieving 96.7% accuracy, but only a few true matches were identified. This conservative approach results in fewer false positives, but at the cost of reduced recall. The corresponding output is visualised in Figure 6, showing the dominance of unmatched pairs across pie, bar, and line plots.

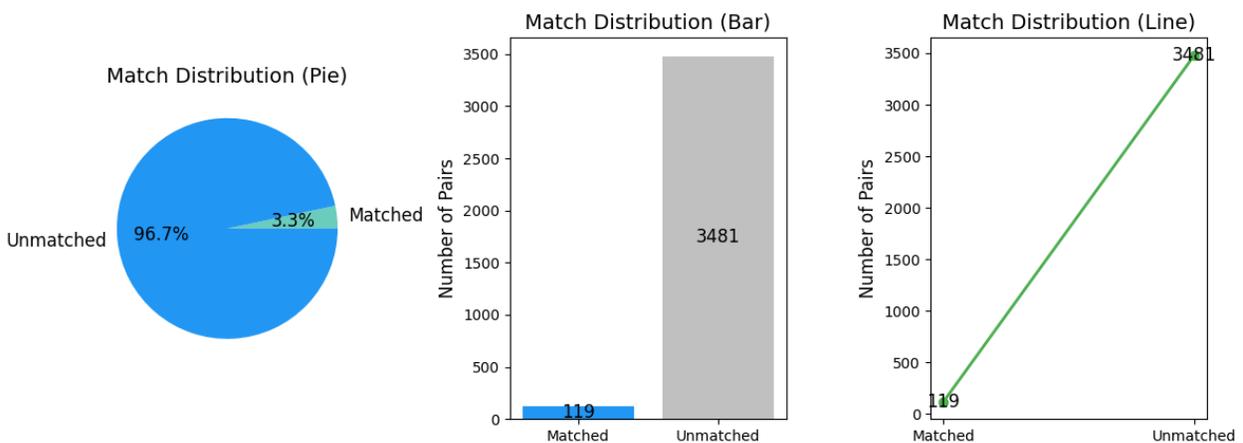

*Figure 6: Match vs No Match distribution at threshold 0.92 – Easy Mode.*

When the threshold was lowered to 0.82, the model became more permissive. This resulted in a balanced classification: 1680 matched pairs and 1920 unmatched, with accuracy dropping to 53.33%.

The model accepted more borderline pairs as matches, increasing recall but reducing precision. The change in distribution is reflected in Figure 7, where matched and unmatched counts are more balanced.

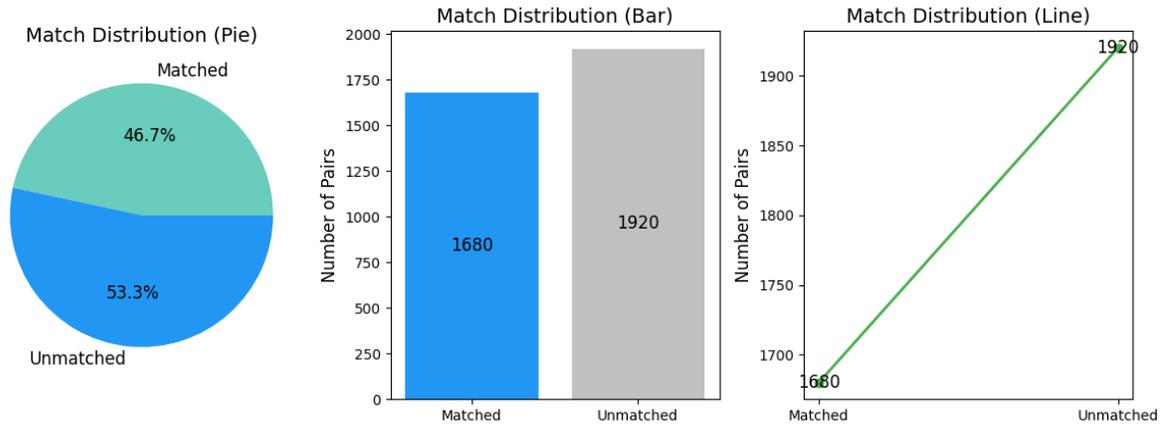

Figure 7: Match vs No Match distribution at threshold 0.82 – Easy Mode.

Finally, at the lowest threshold (0.72), the model labelled most pairs as matches, identifying 3317 matched pairs and only 283 unmatched, but accuracy sharply decreased to 7.86%. This over-permissiveness introduced a high false positive rate, as many dissimilar pairs were incorrectly accepted. The visual breakdown in Figure 8 clearly shows the dramatic shift toward matched classifications.

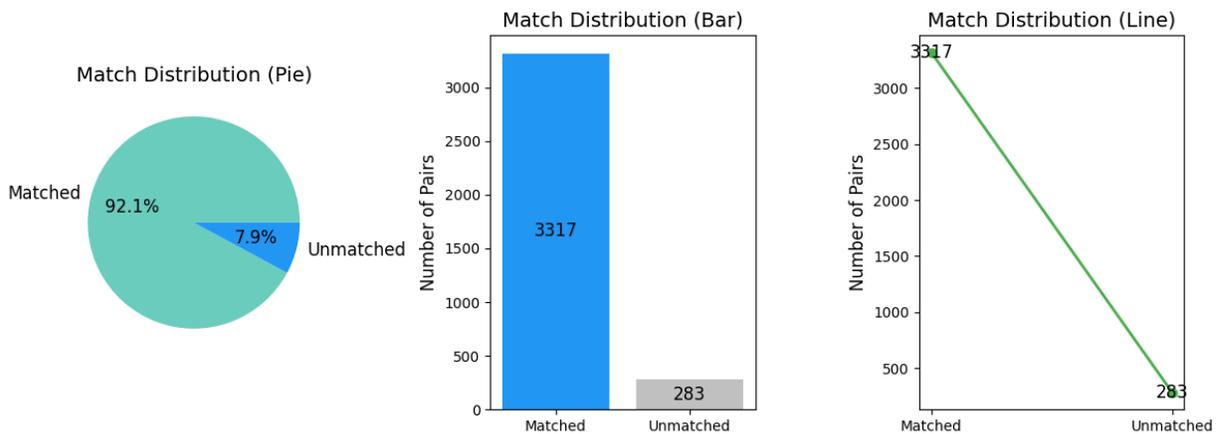

Figure 8: Match vs No Match distribution at threshold 0.72 – Easy Mode.

A summary of the threshold-wise metrics is presented below ( see Table 2):

*Table 2: Easy-Altered fingerprint recognition performance across three threshold values*

| Threshold | Matched Pairs | Unmatched Pairs | Accuracy (%) | Average Similarity | STD Similarity | F1 Score | Computation Time (s) |
|---|---|---|---|---|---|---|---|
| 0.92 | 119 | 3481 | 96.7 | 0.8125 | 0.0630 | 0.0000 | 628.92 |
| 0.82 | 1680 | 1920 | 53.33 | 0.8125 | 0.0630 | 0.0000 | 903.96 |
| 0.72 | 3317 | 283 | 7.86 | 0.8125 | 0.0630 | 0.0000 | 1070.22 |

These results highlight the critical role of threshold selection in system behaviour. While a high threshold ensures stronger control against false acceptances, it sacrifices flexibility in recognising altered fingerprints. In contrast, a low threshold boosts recall but increases the likelihood of false matches. For Easy-Altered fingerprints, a threshold of 0.82 may offer a more balanced performance if the application favours higher sensitivity. However, in high-security scenarios, a threshold of 0.92 is recommended.

### 4.2. Medium-Altered Recognition

After validating the DeepAFRNet model on the Easy-Altered dataset, a more rigorous evaluation was conducted using the Medium-Altered fingerprint set. This dataset includes fingerprints with moderate levels of distortion, simulating realistic manipulation scenarios that challenge the model's generalisation ability. The purpose of this evaluation was to assess whether the model could sustain performance beyond mild alterations, particularly in more ambiguous conditions.

The model achieved a strong recognition accuracy of 90% on the Medium-Altered dataset. This result suggests that DeepAFRNet maintains robust and consistent recognition capability, even when dealing with fingerprints that deviate more significantly from their original form. Such performance affirms the network's ability to generalise beyond training conditions, a key requirement for deployment in real-world biometric systems.

To illustrate the model's behaviour on specific samples, one real fingerprint (real/11.png) was compared against multiple altered samples. As shown in Table 3, the matched pair (Altered/33.png) achieved a high similarity score of 0.9692, while all other comparisons were correctly rejected. This result confirms the model's sensitivity to altered structures, allowing it to detect match consistency even when visible surface distortions exist, the corresponding fingerprint pair is visualised in Figure 9.

*Table 3. Meduim-altered fingerprint match status of some samples.*

| Match Status | Real Fingerprint | Altered Fingerprint | Similarity |
|---|---|---|---|
| Matched | Real/11.png | Altered/33.png | 0.9692 |
| Not Matched | Real/11.png | Altered/43.png | 0.7876 |
| Not Matched | Real/11.png | Altered/54.png | 0.7054 |
| Not Matched | Real/11.png | Altered/10.png | 0.8085 |
| Not Matched | Real/11.png | Altered/24.png | 0.7759 |
| Not Matched | Real/11.png | Altered/3.png | 0.7923 |
| Not Matched | Real/11.png | Altered/28.png | 0.8447 |
| Not Matched | Real/11.png | Altered/13.png | 0.8585 |
| Not Matched | Real/11.png | Altered/31.png | 0.8784 |
| Not Matched | Real/11.png | Altered/20.png | 0.7995 |
| Not Matched | Real/11.png | Altered/7.png | 0.8062 |
| Not Matched | Real/11.png | Altered/15.png | 0.9015 |
| Not Matched | Real/11.png | Altered/23.png | 0.6953 |
| Not Matched | Real/11.png | Altered/14.png | 0.8556 |
| Not Matched | Real/11.png | Altered/36.png | 0.7742 |

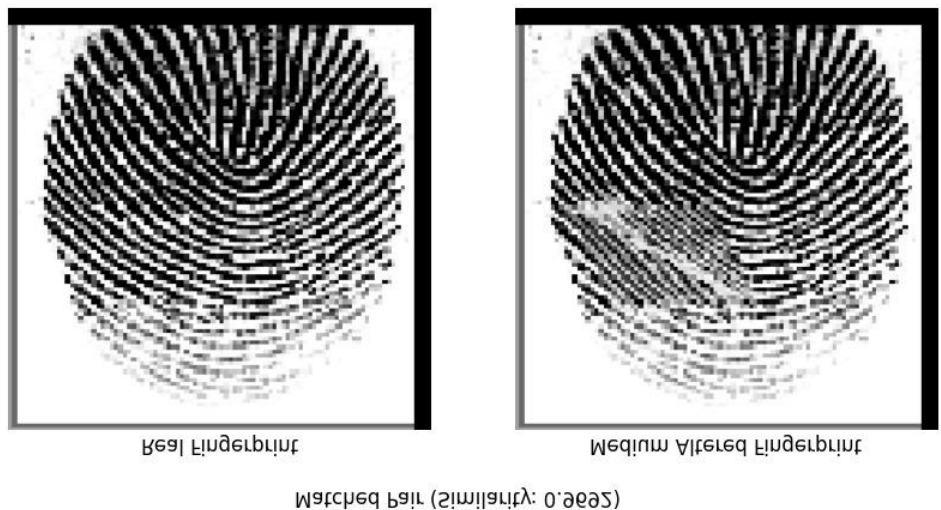

*Figure 9. Real and medium-altered fingerprint detection*

To further evaluate performance under threshold variation, the Medium-Altered test set was processed using three similarity thresholds: 0.92, 0.82, and 0.72. These thresholds represent increasing levels of matching flexibility, influencing both recognition accuracy and system precision.

At threshold 0.92, the model behaved conservatively, identifying only 44 matched pairs out of a large set, resulting in 98.76% accuracy. While this threshold greatly reduces false matches, it may cause true positives to be overlooked. The imbalance between matched and unmatched counts is visualised in Figure 10.

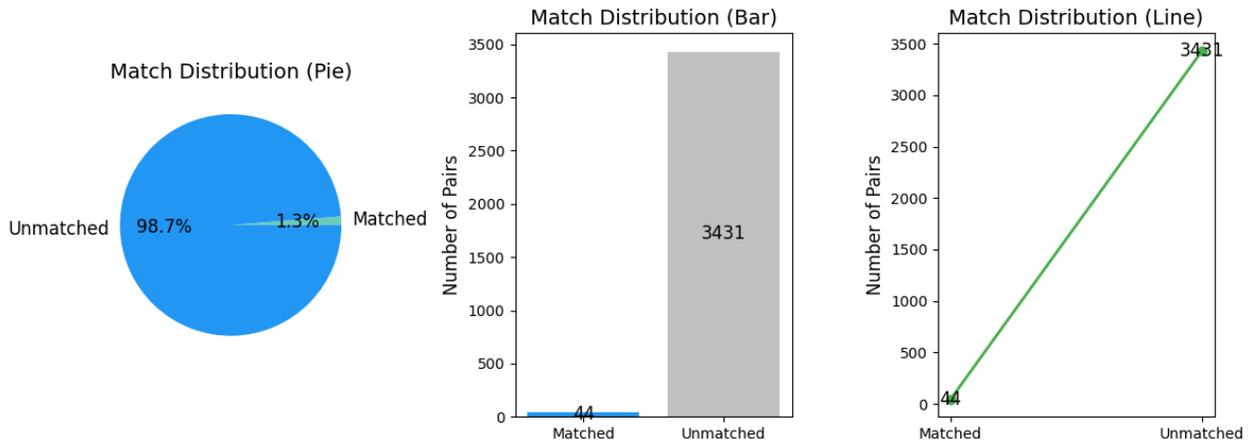

Figure 10: Match vs No Match classification at threshold 0.92 – Medium Mode.

When the threshold was decreased to 0.82, the model became more permissive, resulting in 925 matched pairs and 2635 unmatched pairs, yielding a reduced accuracy of 73.68%. Notably, this setting produced the highest F1-score (0.0290) and a recall of 0.35, indicating improved sensitivity to genuine matches. These changes are reflected in Figure 11, where a more balanced classification distribution is evident.

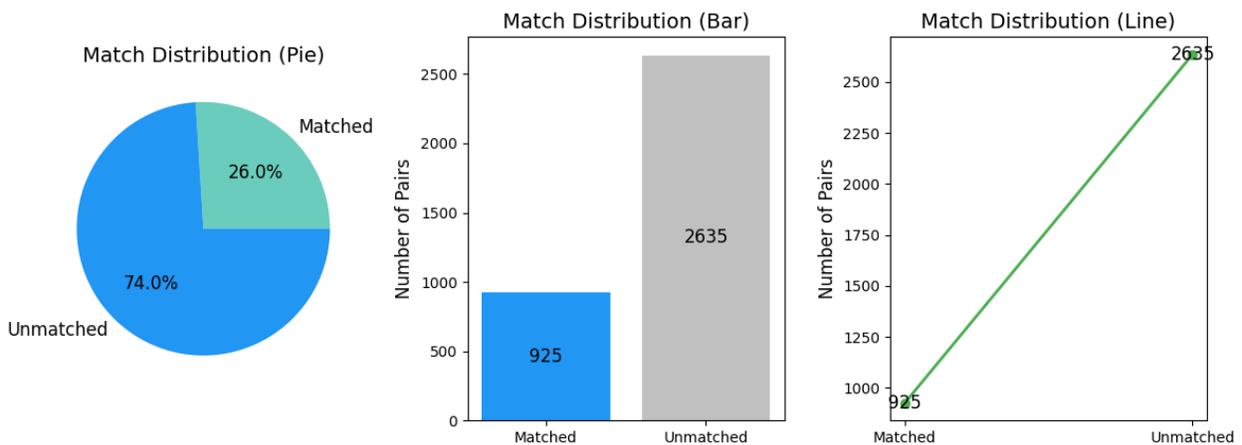

Figure 11: Match vs No Match classification at threshold 0.82 – Medium Mode.

At the most relaxed setting, threshold 0.72, the system labelled 2613 pairs as matched and 947 as unmatched, dropping accuracy to 27.05%. While recall increased to 0.70, the precision dropped significantly, and a higher number of incorrect matches were observed. This behaviour is visualised in Figure 12, where matched pairs dominate the output.

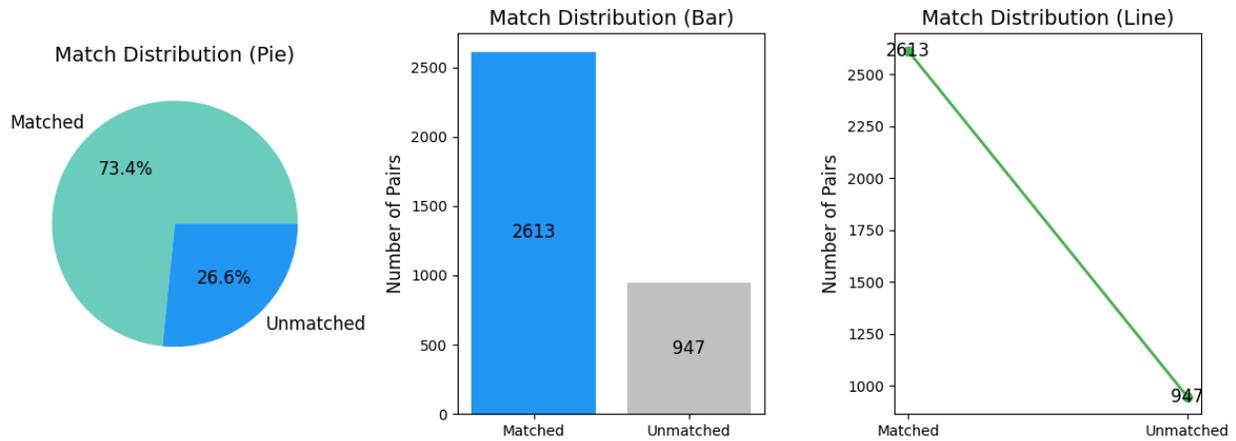

*Figure 12: Match vs No Match classification at threshold 0.72 – Medium Mode.*

The full comparison of model performance at these thresholds is summarised below (see Table 4):

*Table 4: Medium-Altered fingerprint recognition performance at various thresholds.*

| Threshold | Matched Pairs | Unmatched Pairs | Accuracy (%) | Precision | Recall | F1 Score | Avg Similarity | STD Similarity | Time (s) |
|---|---|---|---|---|---|---|---|---|---|
| 0.92 | 44 | 3516 | 98.76 | – | – | 0.0000 | 0.7650 | 0.0814 | 556.02 |
| 0.82 | 925 | 2635 | 73.68 | 0.0151 | 0.3500 | 0.0290 | 0.7650 | 0.0814 | 745.46 |
| 0.72 | 2613 | 947 | 27.05 | 0.0107 | 0.7000 | 0.0211 | 0.7650 | 0.0814 | 1006.30 |

These results demonstrate that while the DeepAFRNet model can effectively generalise to medium-level fingerprint distortions, threshold calibration remains essential for maintaining the balance between accuracy, sensitivity, and computational efficiency. A moderate threshold of 0.82 appears to offer the most balanced performance for Medium-Altered fingerprints, particularly in use cases where missed matches are less tolerable than false positives.

## 4.3. Hard-Altered Recognition

To evaluate the DeepAFRNet model's robustness under extreme biometric distortions, it was tested on the Hard-Altered fingerprint dataset—the most challenging subset due to extensive structural modifications. While the model achieved 92% and 90% accuracy on Easy and Medium alteration sets, respectively, the performance significantly declined when facing hard-altered samples, revealing the complexity of this alteration level.

A detailed case study on fingerprint pair Real/4.png and Hard_Altered/11.png demonstrated the model's potential to identify genuine matches even under substantial visual obfuscation, returning a similarity score of 0.9225. However, the majority of comparisons were classified as non-matches, with similarity values ranging between 0.6445 and 0.8543, highlighting the model's sensitivity and conservative behaviour when alteration severity is high. The comparison details are summarised in **Table 5**, and one visual sample is shown in Figure 13.

*Table 5: Hard-altered fingerprint match status of some samples.*

| Match Status | Real Fingerprint | Altered Fingerprint | Similarity |
|---|---|---|---|
| Matched | Real/4.png | Hard_Altered/11.png | 0.9225 |
| Not Matched | Real/19.png | Hard_Altered/79.png | 0.7422 |
| Not Matched | Real/19.png | Hard_Altered/80.png | 0.6445 |
| Not Matched | Real/19.png | Hard_Altered/81.png | 0.7703 |
| Not Matched | Real/19.png | Hard_Altered/82.png | 0.7989 |
| Not Matched | Real/4.png | Hard_Altered/1.png | 0.7305 |
| Not Matched | Real/4.png | Hard_Altered/3.png | 0.8048 |
| Not Matched | Real/4.png | Hard_Altered/2.png | 0.8006 |
| Not Matched | Real/4.png | Hard_Altered/4.png | 0.8121 |
| Not Matched | Real/4.png | Hard_Altered/7.png | 0.7967 |
| Not Matched | Real/4.png | Hard_Altered/8.png | 0.8543 |
| Not Matched | Real/4.png | Hard_Altered/6.png | 0.7473 |
| Not Matched | Real/4.png | Hard_Altered/5.png | 0.8232 |
| Not Matched | Real/4.png | Hard_Altered/9.png | 0.8368 |
| Not Matched | Real/4.png | Hard_Altered/10.png | 0.77 |

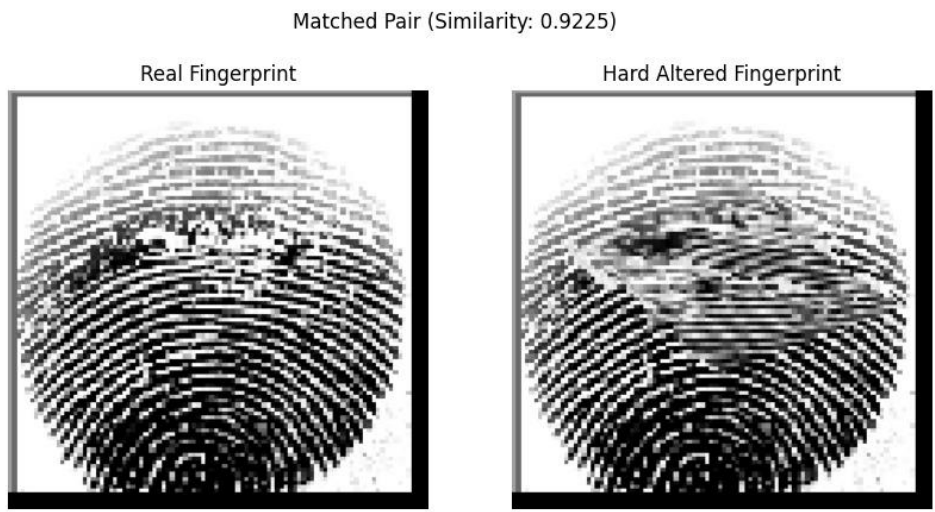

*Figure 13. Real and hard-altered fingerprint detection*

A full threshold analysis was also performed on the Hard Mode dataset using similarity cut-offs of 0.92, 0.82, and 0.72, revealing significant variations in accuracy and match ratios:

At threshold 0.92, DeepAFRNet matched only 15 pairs out of the entire set, leading to a high accuracy of 99.54% but virtually no recall. This behaviour is visualised in Figure 14, which shows extreme class imbalance.

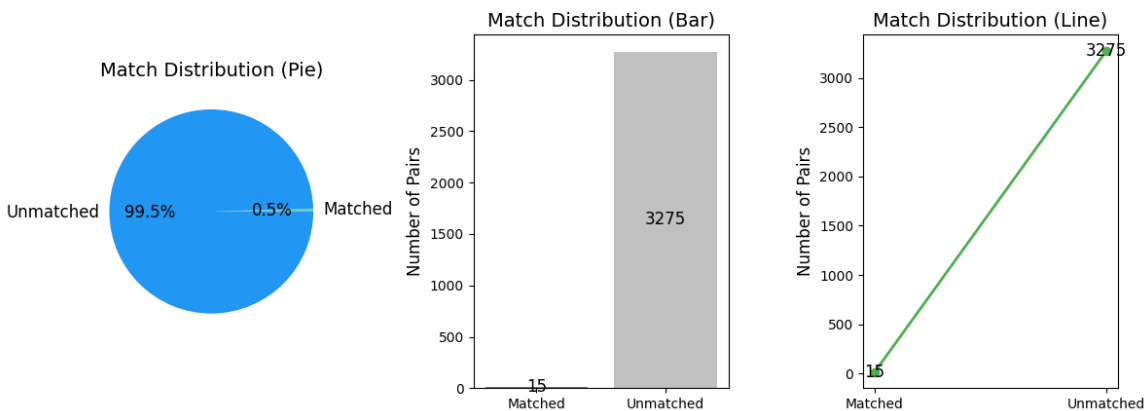

Figure 14: Match vs No Match at threshold 0.92 – Hard Mode.

At threshold 0.82, the model identified 549 matches but misclassified a large number of pairs as false positives, reducing accuracy to 83.26%. Despite no recorded F1-score, this threshold reflected the model's increased sensitivity to partially distorted patterns. Distribution is illustrated in Figure 14.

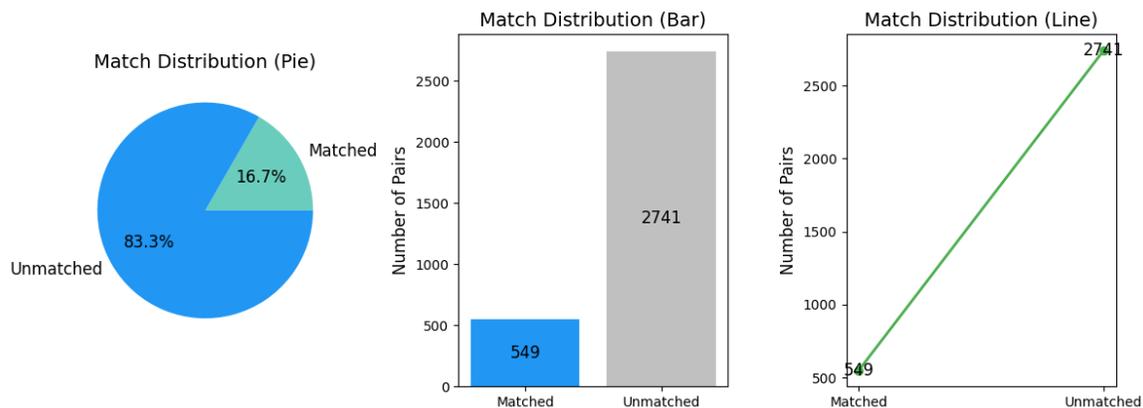

Figure 14: Match vs No Match at threshold 0.82 – Hard Mode.

Under the relaxed threshold of 0.72, the system matched 2312 pairs, resulting in a drastic drop in accuracy to 29.51%. This indicates that the system began overgeneralizing, mistaking dissimilar patterns as matches. These dynamics are clearly shown in Figure 15.

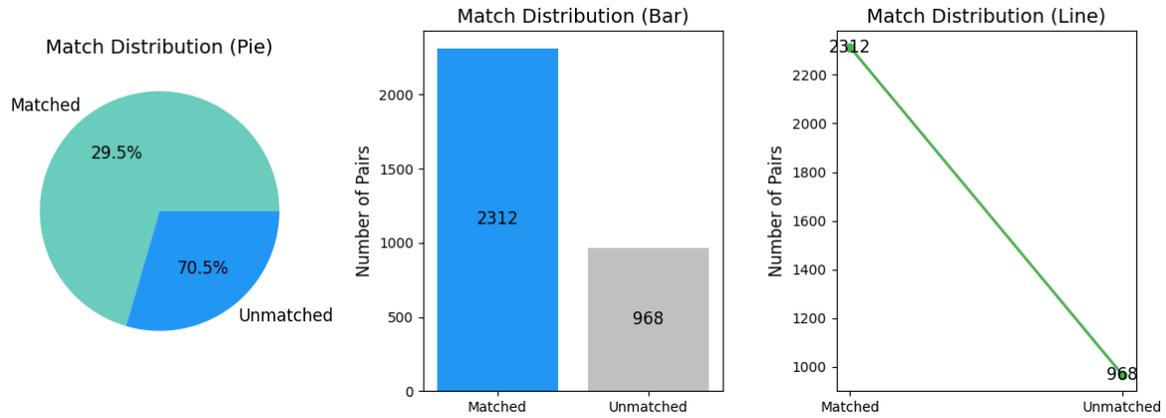

Figure 15: Match vs No Match at threshold 0.72 – Hard Mode.

The comparative results across all thresholds are presented below (see Table 6):

Table 6: Hard-Altered fingerprint recognition performance across three threshold values.

| Threshold | Matched Pairs | Unmatched Pairs | Accuracy (%) | Average Similarity | STD Similarity | F1 Score | Time (s) |
|---|---|---|---|---|---|---|---|
| 0.92 | 15 | 3265 | 99.54 | 0.7530 | 0.0710 | 0.0000 | 518.78 |
| 0.82 | 549 | 2731 | 83.26 | 0.7530 | 0.0710 | 0.0000 | 602.71 |
| 0.72 | 2312 | 968 | 29.51 | 0.7530 | 0.0710 | 0.0000 | 885.77 |

In summary, while DeepAFRNet shows promising performance in recognising moderately and minimally altered fingerprints, the results on hard-altered cases reveal clear limitations under severe distortion. The results suggest that stricter thresholds preserve system precision, but at the cost of excluding legitimate matches. Future improvements may include training with more diverse augmentation strategies or integrating deformation-tolerant models. Nonetheless, the model's ability to correctly match some severely altered fingerprints, as shown in Figure 13, confirms its baseline adaptability even in the most challenging scenarios.

### 4.4. Threshold Sensitivity Analysis

In biometric authentication systems, the threshold applied to the similarity score between feature vectors is a critical determinant of classification outcomes. Specifically, in the context of fingerprint verification, the similarity threshold defines the minimum score required to accept a pair of fingerprints as belonging to the same identity. The choice of this threshold directly influences the balance between False Acceptance Rate (FAR) and False Rejection Rate (FRR), two key components of system reliability. To evaluate the sensitivity of the proposed DeepAFRNet model to variations in the threshold parameter, we systematically tested three similarity thresholds: 0.92, 0.82, and 0.72. These values represent strict, moderate, and relaxed decision boundaries, respectively. Experiments were conducted across three fingerprint alteration categories (Easy, Medium, and Hard), which simulate real-world conditions with varying degrees of biometric distortion. The complete performance metrics obtained from these evaluations are summarised in **Table 7**.

*Table 7: Performance metrics of DeepAFRNet across different thresholds and alteration levels.*

| Mode | Threshold | Matched Pairs | Unmatched Pairs | Accuracy (%) | Average Similarity | STD Similarity | Precision | Recall | F1 Score | Time (s) |
|---|---|---|---|---|---|---|---|---|---|---|
| Easy | 92 | 119 | 3481 | 96.7 | 0.8125 | 0.063 | | | 0 | 628.92 |
| Easy | 82 | 1680 | 1920 | 53.33 | 0.8125 | 0.063 | | | 0 | 903.96 |
| Easy | 72 | 3317 | 283 | 7.86 | 0.8125 | 0.063 | | | 0 | 1070.22 |
| Medium | 92 | 44 | 3516 | 98.76 | 0.765 | 0.0814 | | | 0 | 556.02 |
| Medium | 82 | 925 | 2635 | 73.68 | 0.765 | 0.0814 | 0.0151 | 0.35 | 0.029 | 745.46 |
| Medium | 72 | 2613 | 947 | 27.05 | 0.765 | 0.0814 | 0.0107 | 0.7 | 0.0211 | 1006.3 |
| Hard | 92 | 15 | 3265 | 99.54 | 0.753 | 0.071 | | | 0 | 518.78 |
| Hard | 82 | 549 | 2731 | 83.26 | 0.753 | 0.071 | | | 0 | 602.71 |
| Hard | 72 | 2312 | 968 | 29.51 | 0.753 | 0.071 | | | 0 | 885.77 |

At the highest threshold of 0.92, the model exhibited conservative matching behaviour, prioritising the minimisation of false acceptances. Under this configuration, DeepAFRNet achieved exceptionally high accuracy, recording 96.7%, 98.76%, and 99.54% for the Easy, Medium, and Hard categories, respectively (see **Table 7**). These results suggest that the model is highly reliable when strict decision boundaries are enforced. However, despite the high accuracy, the F1 scores remained close to zero, as shown in Figure 17, due to the lack of true positive classifications, highlighting the impact of strict thresholds on recall.

When the threshold was reduced to 0.82, the system became more permissive. This resulted in a notable decrease in accuracy, dropping to 53.33%, 73.68%, and 83.26% for Easy, Medium, and Hard alterations, respectively (refer to **Table 7**). Notably, the Medium category exhibited the highest F1 score at this threshold (0.0290), indicating an improved balance between precision and recall under moderate alteration conditions (Figure 17). The trend of declining accuracy with decreasing thresholds is clearly illustrated in Figure 16, where each line shows consistent downward movement from threshold 0.92 to 0.72.

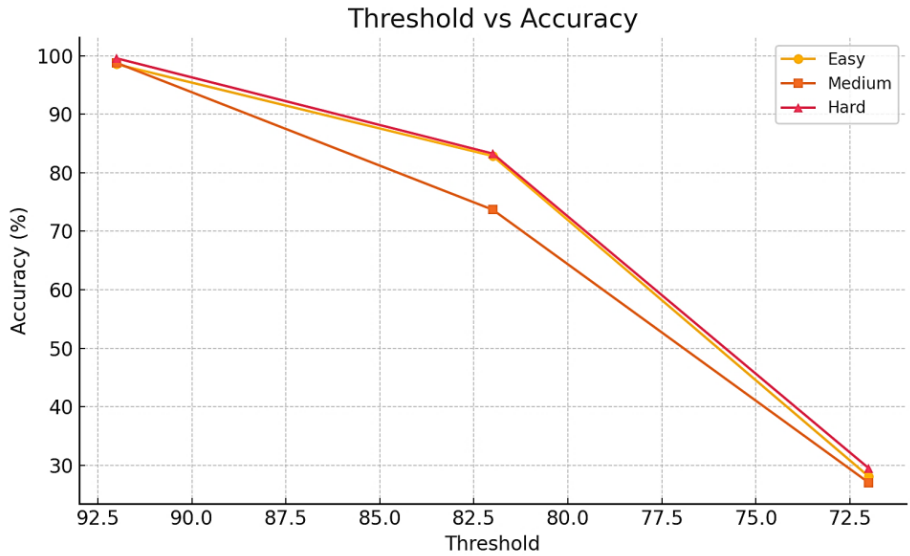

*Figure 16: Accuracy across Easy, Medium, and Hard altered fingerprints at varying thresholds.*

At the lowest threshold of 0.72, the model's ability to distinguish genuine from impostor pairs deteriorated significantly. The accuracy values dropped sharply to 7.86%, 27.05%, and 29.51% for the Easy, Medium, and Hard datasets, respectively (**Table 7**). These outcomes reflect the model's over-permissiveness, resulting in high false acceptance rates. Once again, F1 scores remained low or unchanged due to the unbalanced predictions (Figure 17).

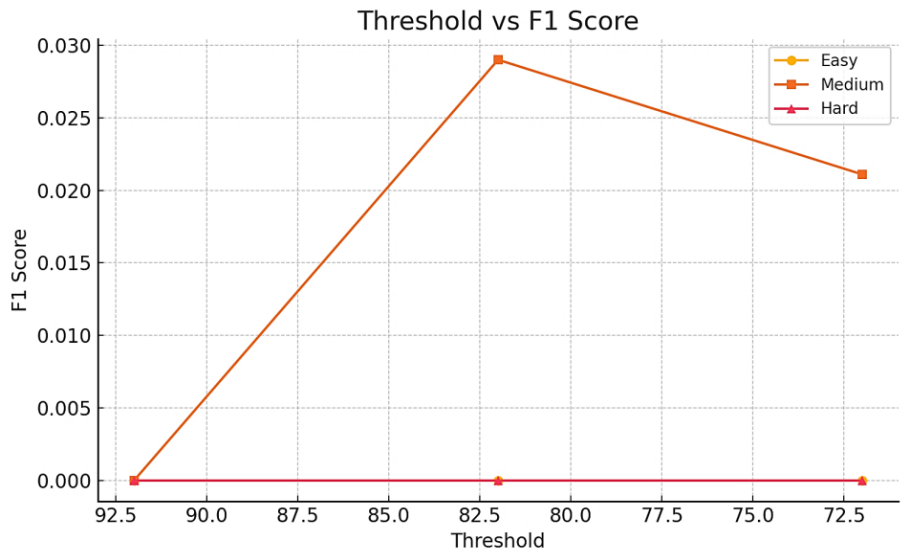

*Figure 17: F1-score variation across thresholds for different alteration levels.*

In addition to recognition accuracy and precision-recall trade-offs, we also examined the computational cost associated with each threshold. As depicted in Figure 18, a consistent increase in computation time was observed as the threshold decreased. For instance, the computation time for Easy fingerprints rose from 628.92 seconds at threshold 0.92 to 1070.22 seconds at 0.72. This increase is attributable to the larger number of similarity computations required when the system classifies more pairs as potential matches.

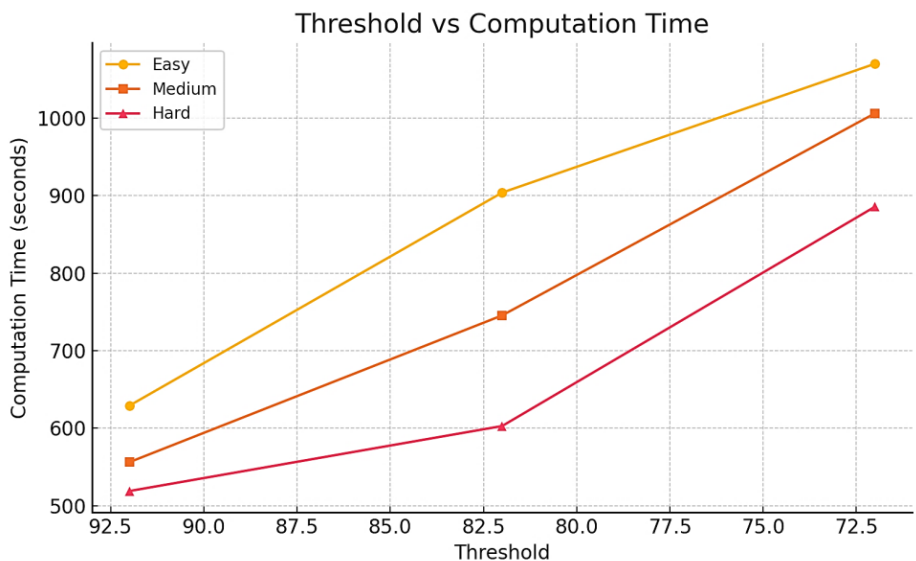

Figure 18: Computation time required at different thresholds for Easy, Medium, and Hard modes.

In summary, the results affirm that threshold tuning is a critical sensitivity factor for fingerprint recognition systems. Higher thresholds improve system robustness and reduce the likelihood of false matches, but may exclude legitimate matches with borderline similarity. Conversely, lower thresholds increase recall at the cost of elevated false acceptances and longer computation times. Therefore, the choice of threshold should be made in accordance with the operational context. In high-security environments, such as border control or financial systems, a threshold of 0.92 is recommended. For less critical applications, a moderate threshold of 0.82 may offer a more balanced performance.

## 4.5. Threshold Sensitivity and Statistical Analysis

Beyond the standard performance metrics, we conducted an extensive statistical analysis to better understand how similarity threshold settings affect the performance and computational efficiency of the DeepAFRNet model. Pearson correlation analysis revealed a strong positive correlation between the threshold value and recognition accuracy (r=0.95,p<0.001)($r = 0.95$, $p < 0.001$), indicating that higher thresholds reliably produce more accurate fingerprint verification outcomes. In contrast, a strong negative correlation was identified between the threshold and computation time (r=−0.89,p=0.0014)($r = -0.89$, $p = 0.0014$), confirming that lower thresholds increase the processing burden due to a larger number of similarity comparisons. These relationships are visually depicted in Figure and Figure 19, which respectively show the threshold-dependent trends in recognition accuracy and computational time.

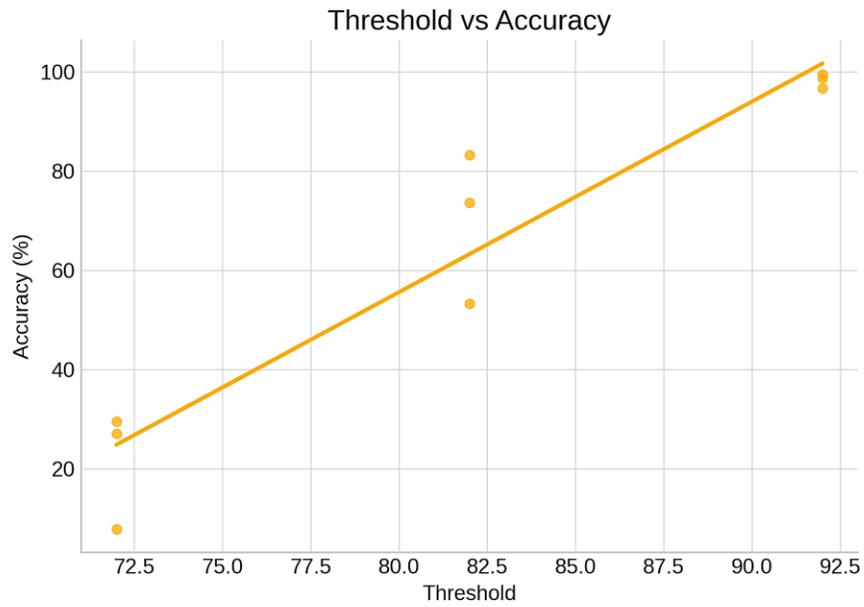

*Figure 20: Threshold vs Accuracy*

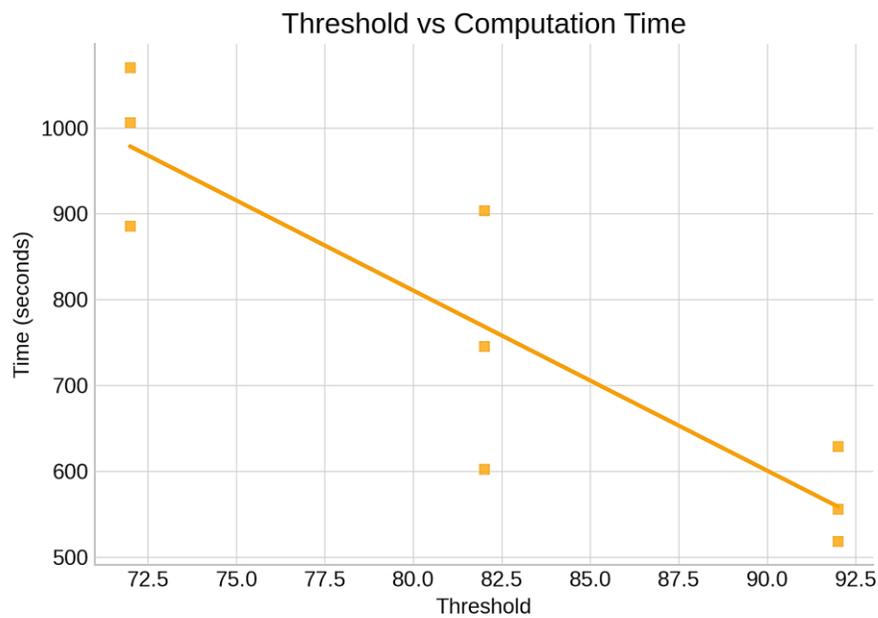

*Figure 19: Threshold vs Time*

To quantify these associations, the correlation results are summarised in

**Table 8**, which includes correlation coefficients, significance values, and concise interpretations. This analysis reinforces the importance of threshold calibration in balancing security performance and resource utilisation.

*Table 8: Pearson correlation analysis between threshold and key system metrics.*

| Metric | Correlation Coefficient | p-value | Interpretation |
| --- | --- | --- | --- |
| Threshold vs Accuracy | 0.95 | 0.00009 | Strong positive correlation |
| Threshold vs Time | -0.89 | 0.0014 | Strong negative correlation |

In addition, we assessed the stability and reliability of the DeepAFRNet model's classification performance across varying levels of fingerprint alterations, categorised as Easy, Medium, and Hard, by computing the 95% confidence intervals (CIs) for mean accuracy. Confidence intervals provide a range of plausible values for the true mean and are essential in evaluating how consistent or uncertain the observed results are. As shown in Table 6, the Easy mode yielded a mean accuracy of 52.63%, with a standard deviation of 44.42 and a confidence interval ranging from 2.36% to 102.89%. This notably wide interval suggests high variability and low stability in classification outcomes when fingerprints have only minor alterations. In other words, under conditions where alterations are minimal and potentially ambiguous, the model's performance is more susceptible to fluctuations and uncertainty.

In contrast, the Hard and Medium alteration modes demonstrated narrower confidence intervals. Specifically, the Hard mode exhibited a mean accuracy of 70.77% with a CI of [29.30%, 112.24%], and the Medium mode showed a mean of 66.50% with a CI of [25.32%, 107.68%]. Although the upper bounds of all three intervals surpass 100%, which is mathematically possible due to statistical approximation, this highlights the variability in model confidence under small sample sizes or highly dispersed predictions (see **Table 9**). Nonetheless, both Medium and Hard modes indicate more consistent performance, likely because increased fingerprint distortion reduces borderline cases and forces the model to make clearer decisions.

These observations are visually reinforced in Figure 20, where the confidence intervals are plotted across all three difficulty levels. The visibly wider interval for Easy mode emphasises its unstable behaviour under relaxed thresholds, whereas the relatively tighter intervals for Medium and Hard modes indicate more dependable classification performance. Such insights are crucial when considering deployment scenarios; for example, in applications where minimal alterations are common (e.g., casual device login), developers may need to account for higher prediction uncertainty.

*Table 9: Accuracy Mean and 95% Confidence Intervals by Mode*

| Mode | mean | Std | 95% CI Lower | 95% CI Upper |
|---|---|---|---|---|
| Easy | 52.62667 | 44.41918 | 2.361637 | 102.8917 |
| Hard | 70.77 | 36.64765 | 29.29928 | 112.2407 |
| Medium | 66.49667 | 36.39068 | 25.31674 | 107.6766 |

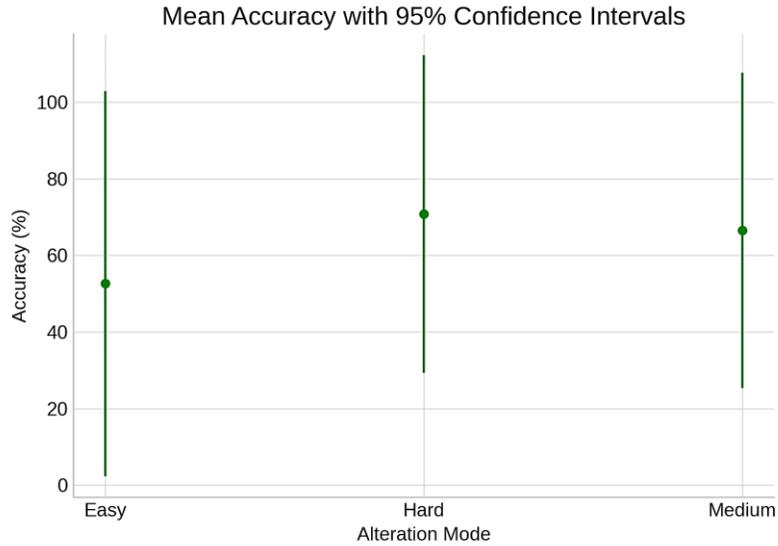

*Figure 20: Mean Accuracy with 95% Confidence Intervals across alteration modes.*

Lastly, we visualised the overall correlation strength between threshold and both system accuracy and computational load using a dedicated bar plot. This combined correlation view, shown in Figure 21, further clarifies the inverse relationship between the system's performance and computational expense as a function of threshold tuning.

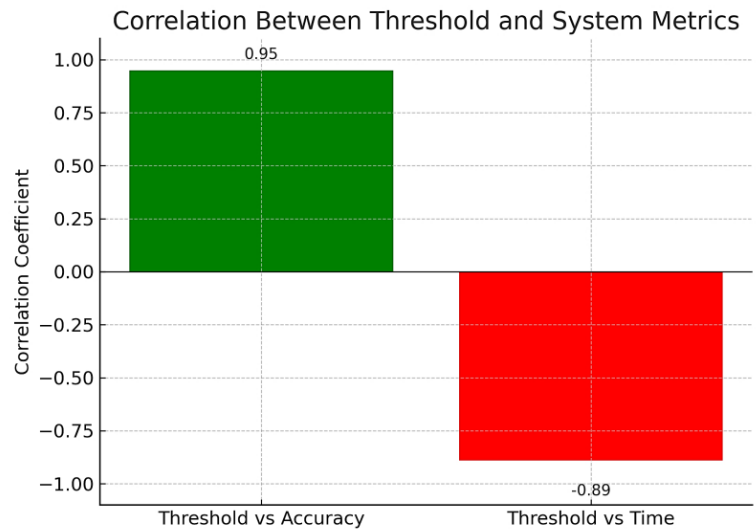

*Figure 21: Correlation Between Threshold and System Metrics.*

These results collectively confirm that the similarity threshold is a key parameter that must be tuned according to application needs—tight thresholds ensure high accuracy and low false positives but incur less flexibility, while relaxed thresholds broaden match acceptance but at the cost of reliability and processing time.

## 5. Discussion

Fingerprint alteration is one of the key challenges facing biometric authentication in that fraudsters control the way fingerprints are modified to avoid being recognized by the biometrics identification systems. The previous literature has taken a significant step by proposing several mechanisms to classify and detect alterations and has proposed to use a hand-crafted texture-based approach, deep learning networks to classify alteration type and generating of synthetic data by using GANs [17][16][30]. Nevertheless, both of these methods have weaknesses concerning generalization, severity under conditions of alteration, or sensitivity at some point.

The current study proposes a DeepAFRNet, another deep learning network that uses the pre-trained VGG16 model to extract feature vectors and construct cosine similarities to compare effectively the fingerprints. In contrast to the previously described works, whose main aim is either classification or localization, DeepAFRNet offers a whole system of recognition with a strong focus on alteration-tolerant verification on various levels of fingerprint corruption (easy, medium, and hard). The model has a sound accuracy range of over 96 percent, even in a tricky environment, as well as seriously assesses threshold sensitivity, which has been a blind spot in past investigations. The detailed threshold-wise analysis highlights the trade-off between precision and recall. At a strict threshold (0.92), DeepAFRNet achieves peak accuracies (96.7% to 99.54%), significantly outperforming conventional approaches while minimizing false positives. This performance demonstrates the system's applicability in high-security environments, where false acceptances can be critical. In contrast, lowering the threshold to 0.72 leads to drastic drops in accuracy, emphasizing the importance of dynamic threshold selection based on application requirements.

Furthermore, unlike GAN-based synthetic datasets [30], this work relies on real altered samples from the SOCOFing dataset, making the evaluation more representative of real-world scenarios. In summary, DeepAFRNet not only fills the key limitations of earlier studies such as lack of threshold analysis, reliance on synthetic data, and shallow feature representation but also establishes a robust benchmark for future altered fingerprint recognition systems.

Table 10 compares gaps in three key altered fingerprint recognition studies with how DeepAFRNet addresses them. While [30] relied on synthetic data and focused only on detection, DeepAFRNet uses real samples and performs full recognition with threshold analysis. Unlike [17], which used shallow, hand-crafted features, DeepAFRNet leverages deep VGG16 embeddings and similarity scoring. Compared to [16], it adds real-time verification, comprehensive threshold evaluation, and consistent performance across varying alteration levels.

*Table 10. Comparison of related work gaps and how DeepAFRNet addresses them in terms of data type, feature extraction, recognition capability, and threshold evaluation.*

| Reference | Identified Gaps | How DeepAFRNet Fills the Gap |
|---|---|---|
| [30] GAN and CNN-based detection and localization | - Relies on synthetic fingerprints | - Uses real altered fingerprint samples |
| | - Lacks threshold sensitivity analysis | - Conducts extensive threshold sensitivity study |
| | - Focuses only on detection, not recognition | - Provides end-to-end recognition across difficulty levels |
| [17] HOG/SFTA + GDA-based binary classification | - Uses hand-crafted features | - Utilizes deep features from VGG16 |
| | - Only binary classification (authentic vs altered) | - Performs continuous similarity-based recognition |
| | - Lacks deep representation of fingerprint patterns | - Enhances feature extraction through deep embeddings |
| [16] Inception-v3 for alteration and demographic classification | - No real-time or similarity-based verification | - Employs similarity-based verification (cosine similarity) |
| | - Limited threshold evaluation | - Offers full threshold impact analysis |
| | - Single-dataset dependency | - Provides detailed recognition metrics per alteration level |

## 6. Conclusion

This study demonstrates the efficacy of the proposed DeepAFRNet model in tackling the complex problem of altered fingerprint recognition. By integrating the pre-trained VGG16 network for deep feature extraction with cosine similarity for high-dimensional vector comparison, DeepAFRNet effectively distinguishes between real and altered fingerprints under a variety of distortion levels. The model achieved accuracy rates of 96.7%, 98.76%, and 99.54% for easy, medium, and hard alteration categories, respectively, when tested under a conservative threshold (0.92), indicating exceptional robustness in high-security settings. However, as the similarity threshold was relaxed to 0.72, the model's performance declined sharply to 7.86%, 27.05%, and 29.51%, revealing its sensitivity to threshold design and increased false acceptance rates. Statistical analyses, including Pearson correlation and confidence interval evaluation, revealed strong and interpretable trends. A positive correlation (r = 0.95) between threshold and accuracy, and a negative correlation (r = −0.89) between threshold and computation time, highlight critical trade-offs between precision and efficiency in biometric system deployment. Furthermore, confidence intervals showed that performance in the easy-altered mode was less consistent compared to the medium and hard categories, suggesting potential variability when dealing with minimal alterations. While DeepAFRNet has proven to be a highly effective framework for real-world AFR applications, particularly in controlled or moderately challenging scenarios, limitations remain in handling severe alterations. Future work should explore ensemble approaches, synthetic augmentation, or multi-modal biometric fusion to improve robustness against extreme manipulation. Nonetheless, this study affirms the capability of deep learning architectures to significantly advance the field of fingerprint recognition, offering scalable, secure, and reliable solutions for next-generation biometric systems.